\documentclass[11pt]{article}

\usepackage[preprint]{acl}

\usepackage{times}
\usepackage{latexsym}

\usepackage[T1]{fontenc}

\usepackage[utf8]{inputenc}

\usepackage{microtype}
\usepackage{multirow}
\usepackage{inconsolata}

\usepackage{graphicx}

\title{Confidence, Not Perplexity: A Better Metric for the Creative Era of LLMs}

\author{V. S. Raghu Parupudi\\
  University of California, San Diego \\
  La Jolla, CA 92092 \\
  \texttt{pvsrrkishore@gmail.com} \\
  \texttt{s1parupudi@ucsd.edu} \\}

\begin{document}
\maketitle
\begin{abstract}

Reference-free metrics like self-perplexity are strongly biased against creative text generation. We propose the Confidence Score (CS), derived from a model's output probability distribution, as a less biased alternative. Experiments on gpt-4o-mini show that while fluency-based metrics prefer novel responses in 0\% of cases on 99 creative prompts, our CS does so 19\% of the time—a statistically significant difference (95\% CI for difference: [11.1\%, 27.3\%]). We also show that CS effectively distinguishes between easy, medium, and hard tasks, confirmed by non-overlapping confidence intervals. The Confidence Score thus mitigates the creativity bias of traditional metrics while retaining their core evaluative strengths, offering a more balanced assessment for modern LLMs.
\end{abstract}

\section{Introduction}

The rapid advancement of Large Language Models (LLMs) has outpaced the development of robust evaluation methods, particularly for open-ended creative tasks where human references are impractical. The standard reference-free metric, self-perplexity \citep{shannon1948, brown1993perplexity}, is increasingly questioned for this purpose. Unlike reference-based metrics such as BLEU \citep{papineni2002bleu}, perplexity can operate without a ground truth, but it has a fundamental flaw: it rewards models for making safe, high-frequency word choices, which leads to dull, repetitive "neural text degeneration" \citep{holtzman2020curious} and correlates poorly with human quality judgments \citep{fabbri2021summeval}. While modern metrics have moved towards semantic similarity \citep{zhang2020bertscore}, they do not directly probe the model's internal state at the moment of generation. To address this gap, we propose the Confidence Score (CS), a novel reference-free metric that analyzes the statistical properties of the output probability distribution to quantify model certainty. We validate CS through two experiments, demonstrating that it is significantly less biased against creative, high-temperature responses than perplexity, while remaining as effective at distinguishing between tasks of varying difficulty. Our findings present a more balanced and robust evaluation framework for the creative era of LLMs.

\section{Related Work}

he evaluation of machine-generated text has evolved alongside model capabilities. The traditional approach is rooted in information theory \citep{shannon1948}, with perplexity serving as the standard for decades for quantifying model uncertainty \citep{brown1993perplexity}. Its ability to function without a reference text made it a practical choice, in contrast to reference-based n-gram overlap metrics like BLEU \citep{papineni2002bleu}. However, the limitations of these classic metrics are now well-documented. A seminal work by \citet{holtzman2020curious} demonstrated that optimizing for low perplexity leads to "neural text degeneration"—outputs that are bland and repetitive. This finding highlights a fundamental bias in fluency-based scores towards high-frequency, predictable language, which helps explain their consistently low correlation with human judgments of quality \citep{fabbri2021summeval}. These critiques motivate the need for metrics that do not equate predictability with quality.

Recent research has pursued several new paradigms for reference-free evaluation. One direction has been to move beyond lexical overlap to semantic similarity, with metrics like BERTScore \citep{zhang2020bertscore} using contextual embeddings to offer a richer sense of meaning. Another popular approach is the "LLM-as-a-judge" method, where a powerful model like GPT-4 is prompted to score the outputs of other models \citep{liu2023gevalnlgevaluationusing}. While often correlating well with human ratings, this approach can be computationally expensive and lacks transparency. A third, emerging direction leverages the generative model's own uncertainty. For instance, Self-Check GPT \citep{manakul2023selfcheckgpt} probes an LLM's uncertainty by sampling multiple responses to detect potential hallucinations. Our work is most closely related to this third paradigm. However, where Self-Check GPT requires multiple generation passes, our proposed Confidence Score (CS) directly quantifies the model's certainty from the probability distribution of a single generation, offering a more computationally efficient and direct measure of confidence.

\section{The Confidence Score Metric}
\subsection{The Core Limitation of Perplexity}

At its core, self-perplexity is a measure of a model's "surprise" at its own output, calculated as the exponentiated average negative log-probability of the generated sequence. For a single token $w_{t}$ generated at timestep t given the context $c_{t}$, the score is derived from its conditional probability, P($w_{t}$|$c_{t}$). This formulation, while mathematically simple, reveals a critical limitation: it only considers the probability of the single token that was chosen, ignoring the probabilities of all the tokens that were not chosen.

This makes perplexity an unreliable indicator of a model's true "confidence." For example, a model may assign a probability of 0.4 to its chosen token in two different scenarios: one where the next most likely token has a probability of 0.39 (high uncertainty), and another where the next most likely token has a probability of 0.1 (high certainty). Self-perplexity would assign the same score to both of these tokens, failing to distinguish between a confident decision and a guess. This flaw makes it susceptible to favoring fluent but simplistic outputs composed of high-frequency words, as these often have high top-1 probabilities even if the model is not decisively certain.

\subsection{Formulating the Confidence Score (CS)}
To address this limitation, we hypothesize that a more robust metric must analyze the shape of the output probability distribution. A truly confident prediction is characterized not just by a high probability for the chosen token, but also by a large gap between the chosen token and its closest alternatives.

We propose the Confidence Score (CS), a metric designed to capture this relationship. The CS for a single token is defined as the product of two key properties of the distribution: the Probability of the Chosen Token ($P_{chosen}$), which serves as the base score representing the raw likelihood of the model's selection, and the Standard Deviation of the Top-N Probabilities ($\sigma _{top_n}$). This second component measures the spread, or dispersion, of probabilities among the top candidates, where a high standard deviation indicates that the top choice is a clear outlier and decisively distinguished from competing tokens.

Formally, let $w_{t}$  be the token generated at timestep t. Let $S_{t}$ be the set of probabilities for the top n most likely tokens at that step, where P($w_{t}$) is the highest probability in the set. The Confidence Score for the token, $CS_{t}$ is:

\begin{equation}
\label{eq:cs}
CS_t = P(w_t) \times \sigma(S_t)
\end{equation}

By multiplying these two values, the CS rewards distributions that are sharply "peaked" around a single dominant choice. A high score is only achieved if the model is both highly certain of its choice (high P($w_{t}$)) and faces little competition from alternatives (high $\sigma(S_t)$).

\subsection{Aggregation Methods}
A single token's score is informative, but the quality of a sequence depends on the aggregate confidence across all its tokens. We therefore propose two methods for aggregating the token-level scores $CS_{t}$ into a final sequence score:

Average CS (Avg. CS): The arithmetic mean of the CS scores for all tokens in the sequence. This provides a holistic measure of the model's overall confidence and is useful for gauging the general quality of an output.

Worst-Case CS (WC CS): The minimum CS score across all tokens in the sequence. This metric acts as a "weakest link" detector, identifying the single point of highest uncertainty in a generation. It is particularly effective at flagging potential factual errors or logical inconsistencies, which often manifest as moments of low model confidence.

By using these two complementary scores, we can develop a more nuanced understanding of a model's output quality.

\section{Experimental Setup}

To empirically validate the properties of our proposed Confidence Score (CS) and compare it against traditional fluency metrics, we designed two distinct experiments. The first aims to quantify the bias of each metric in evaluating creative text, while the second assesses their ability to distinguish between tasks of varying difficulty. All experiments were conducted using the gpt-4o-mini model via the OpenAI API.

\subsection{Datasets}
We constructed four distinct sets of English-language prompts, each designed to elicit a specific type of response from the language model. To ensure the integrity of each category, all questions were manually reviewed and validated by human annotators to confirm they aligned with their intended category's objective.

1. Creative Prompts: This set consists of 100 questions designed to be abstract, philosophical, subjective, or imaginative, thereby encouraging the model to generate novel and diverse responses. These prompts are central to our first experiment on creativity bias. An example from this set is: "Describe the color blue to someone who is blind."

2. Easy Prompts: This set contains 30 unambiguous factual questions with a single, well-known correct answer. The purpose of this set is to establish a baseline for high-confidence, factual recall tasks where the model's uncertainty should be minimal. An example is: "What is the chemical symbol for gold?"

3. Medium Prompts: This set includes 30 questions that require simple reasoning, synthesis of common knowledge, or short explanations, rather than direct factual recall. These prompts are designed to elicit responses that are more complex than the easy tasks but still have a relatively constrained set of correct answers. An example is: "Explain in one sentence why the sky is blue."

4. Hard Prompts: This set contains 30 questions that involve nuanced reasoning, ethical judgment, or complex synthesis. These prompts are open-ended and do not have a single correct answer, challenging the model to construct a coherent and well-reasoned argument. An example is:"What is a compelling argument for and against the universal adoption of AI?"

\subsection{Metric Implementation Details}
For every token generated in our experiments, we requested and stored the top 10 log probabilities from the gpt-4o-mini API. This raw probability data formed the basis for all subsequent metric calculations.

For our Confidence Score (CS) metric, the parameter n (the number of top token probabilities to consider) was set to 3. This value was chosen to capture the most critical local competition dynamics—the relationship between the chosen token and its two closest rivals—without diluting the signal with the long tail of very low-probability tokens from the rest of the distribution.

The standard deviation component of the CS was calculated using the population standard deviation. This choice is deliberate, the set of top-N probabilities is not treated as a sample from which to estimate the properties of the entire vocabulary. It is treated as the complete population of interest as we are directly measuring the dispersion within this specific set of top candidates.

\section{Results}

Our experiments yield two primary findings. First, we provide statistical evidence that our proposed Confidence Score (CS) metrics are significantly less biased against creative text than traditional fluency-based metrics. Second, we demonstrate that our CS metrics are as effective as self-perplexity at the traditional task of distinguishing between outputs generated for tasks of varying difficulty.

\subsection{Experiment 1: Quantifying the Bias Against Creativity}
To assess bias, we compared each metric's preference for low-temperature (T=0.1, stable) versus high-temperature (T=1.1, creative) responses on our set of 99 creative prompts. A metric is considered to prefer the high-temperature response if it assigns it a better score (lower for perplexity, higher for all others).

The results, summarized in Table~\ref{tab:preference_rates}, show a stark contrast. Both Self-Perplexity and the Fluency Score exhibit an extreme bias, preferring the more stable, low-temperature response in 100\% of cases. In contrast, our proposed Average CS and Worst-Case CS metrics show a more balanced evaluation, preferring the creative response 19.2

To determine if this difference is statistically significant, we performed a bootstrapped analysis of the difference in these preference rates. As shown in Table~\ref{tab:preference_diff}, the 95\% confidence intervals for the difference between our CS metrics and both Self-Perplexity and Fluency Score are entirely above zero. This provides strong statistical evidence that both Average CS and Worst-Case CS are significantly less biased against creative text generation than traditional fluency-based metrics. A Wilcoxon signed-rank test confirmed that the score distributions between low- and high-temperature responses were significantly different for Self-Perplexity, Fluency Score, and Average CS (p < 0.001).

\begin{table}[h!]
\centering
\begin{tabular}{lc}
\hline
\textbf{Metric} & \textbf{Preference Rate (95\% CI)} \\
\hline
Perplexity & 0.0\% (0.0\%, 0.0\%) \\
Fluency    & 0.0\% (0.0\%, 0.0\%) \\
Avg. CS    & 19.2\% (11.1\%, 27.3\%) \\
WC CS      & 17.2\% (10.1\%, 25.3\%) \\
\hline
\end{tabular}
\caption{Preference rates for high-temperature (T=1.1) responses on 99 creative prompts. The observed rate is shown with its bootstrapped 95\% confidence interval.}
\label{tab:preference_rates}
\end{table}

\begin{table}[h!]
\centering
\setlength{\tabcolsep}{3pt} 
\begin{tabular}{lc}
\hline
\textbf{Comparison} & \textbf{Mean Diff. (95\% CI)} \\
\hline
Avg. CS vs PPL & +19.2\% (11.1\%, 27.3\%) \\
WC CS vs PPL   & +17.2\% (10.1\%, 25.3\%) \\
\hline
Avg. CS vs Fluency & +19.2\% (12.1\%, 27.3\%) \\
WC CS vs Fluency   & +17.2\% (10.1\%, 25.3\%) \\
\hline
\end{tabular}
\caption{Bootstrapped difference in preference rates for creative responses. Since all CIs are above zero, our CS metrics are significantly less biased.}
\label{tab:preference_diff}
\end{table}

\subsection{Experiment 2: Distinguishing Task Difficulty}

To ensure our metrics are robust for general evaluation, we tested their ability to differentiate between easy, medium, and hard tasks. We generated a single response (T=0.5) for 30 prompts in each category and calculated the 95\% confidence intervals for the mean score of each metric.

The results are presented in Table~\ref{tab:difficulty_results}. For all metrics, the 95\% confidence intervals for the three difficulty categories are mutually exclusive, indicating a statistically significant difference between the groups. This demonstrates that our proposed Average CS and Worst-Case CS are as effective as traditional fluency metrics at distinguishing between task difficulties.

\begin{table}[t!]
\centering
\setlength{\tabcolsep}{2pt}
\begin{tabular}{llc}
\hline
\textbf{Category} & \textbf{Metric} & \textbf{Mean (95\% CI)} \\
\hline
\multirow{4}{*}{Easy} & Perplexity & 1.058 (1.036, 1.084) \\
& Fluency    & -0.055 (-0.075, -0.037) \\
& Avg. CS    & 0.434 (0.423, 0.446) \\
& WC CS      & 0.188 (0.120, 0.258) \\
\hline
\multirow{4}{*}{Medium} & Perplexity & 1.193 (1.168, 1.218) \\
& Fluency    & -0.175 (-0.198, -0.155) \\
& Avg. CS    & 0.364 (0.355, 0.373) \\
& WC CS      & 0.003 (0.001, 0.004) \\
\hline
\multirow{4}{*}{Hard} & Perplexity & 1.267 (1.239, 1.296) \\
& Fluency    & -0.234 (-0.257, -0.211) \\
& Avg. CS    & 0.338 (0.329, 0.350) \\
& WC CS      & 0.002 (0.001, 0.003) \\
\hline
\end{tabular}
\caption{Mean scores and bootstrapped 95\% confidence intervals for each metric across three task difficulties, rounded to three decimal places. The non-overlapping CIs for each metric demonstrate a statistically significant ability to distinguish between categories.}
\label{tab:difficulty_results}
\end{table}

\section{Conclusion}

In this work, we addressed the bias of traditional reference-free metrics, such as self-perplexity and fluency scores, which are ill-suited for evaluating the creative and novel outputs of modern LLMs. We proposed a new metric, the Confidence Score (CS), which moves beyond rewarding mere predictability by analyzing the statistical shape of the model's output probability distribution to measure its certainty.

Our experiments provide strong, statistically significant evidence for the utility of our proposed metrics. We demonstrated that our CS metrics are less biased against creativity compared to both self-perplexity and fluency scores. Then, we showed that the CS metrics are as effective as traditional metrics at the standard task of distinguishing between prompts of varying difficulty.

\section{Limitations and Ongoing Work}

Our current study has several limitations that directly inform our ongoing research. Our analysis is based on a single model, gpt-4o-mini, and we use temperature as a proxy for creativity, which is a proxy for creativity. These limitations are being actively addressed. Ongoing work involves validating the Confidence Score (CS) across different LLMs to establish generalizability. Next, we want to get a large-scale correlation study between CS and direct human judgments of text quality and creativity. We are exploring applications beyond evaluation, with investigations into integrating the token-level Confidence Score directly into decoding algorithms. The goal is to use CS as a real-time signal to mitigate hallucinations, thereby improving the reliability of generative models.

\bibliography{custom}

\end{document}